\pdfoutput=1

\documentclass[11pt]{article}

\usepackage{acl}

\usepackage{times}
\usepackage{latexsym}
\usepackage{graphicx}

\usepackage[T1]{fontenc}

\usepackage[utf8]{inputenc}

\usepackage{microtype}

\usepackage{inconsolata}

\newcommand{\babel}{\textsc{Babel Briefings}}

\title{A diverse Multilingual News Headlines Dataset from around the World}

\author{Felix Leeb \and Bernhard Schölkopf \\
  Max Planck Institute for Intelligent Systems \\
  Tübingen, Germany \\
  \texttt{fleeb@tue.mpg.de} 
}

\begin{document}
\maketitle
\begin{abstract}

\babel{} is a novel dataset featuring 4.7 million news headlines from August 2020 to November 2021, across 30 languages and 54 locations worldwide with English translations of all articles included. Designed for natural language processing and media studies, it serves as a high-quality dataset for training or evaluating language models as well as offering a simple, accessible collection of articles, for example, to analyze global news coverage and cultural narratives. 
As a simple demonstration of the analyses facilitated by this dataset, we use a basic procedure using a TF-IDF weighted similarity metric to group articles into clusters about the same event. We then visualize the \emph{event signatures} of the event showing articles of which languages appear over time, revealing intuitive features based on the proximity of the event and unexpectedness of the event. 
The dataset is available on \href{https://www.kaggle.com/datasets/felixludos/babel-briefings}{Kaggle} and \href{https://huggingface.co/datasets/felixludos/babel-briefings}{HuggingFace} with accompanying \href{https://github.com/felixludos/babel-briefings}{GitHub} code.

\end{abstract}

\section{Introduction}

Analyzing news headlines can be an invaluable source of data for a wide variety of natural language processing tasks such as bias detection~\citep{gangula2019}, topic classification~\citep{rana2014news}, or event tracking~\citep{qian2019detecting}.
Furthermore, news headlines can provide insights for sociologists and political scientists about how people think about and discuss current events.

The coverage and discussion of current events varies significantly across different media outlets worldwide, however, these distinctions may be difficult to integrate in data mining or machine learning systems due to the language barrier. 
There are relatively few datasets offering extensive, diverse, and multilingual content~\citep{kreutzer2022quality}.
This is especially problematic for natural language processing tasks, which have been shown to exhibit language biases~\citep{gallegos2023bias}. 


We seek to address these limitations with a new dataset called \babel{}, which is an accessible dataset representing a wide variety of languages and cultures. \babel{} provides daily headlines of articles from across the world, originally written in one of 30 languages from 54 locations around the world published between August 2020 and November 2021, for a total of about 4.7 million distinct articles. Consequently, our dataset offers a rich source for analyses of world events, cultural narratives, media framing, and more.




We make this dataset available on \href{https://www.kaggle.com/datasets/felixludos/babel-briefings}{Kaggle} and \href{https://huggingface.co/datasets/felixludos/babel-briefings}{HuggingFace} for easy and open access under the CC BY-NC-SA 4.0 license \footnote{This license permits non-commercial use as long as the dataset is credited and variants are licensed under the same terms.}, as well as providing all code used to collect and process the data on \href{https://github.com/felixludos/babel-briefings}{GitHub}.

\subsection{Related Work}

Many comparable datasets focus either on \textit{depth}, i.e., tracking a small number (or even a single) outlet over some time, or \textit{breadth} for comparative studies of specific events. Meanwhile, our dataset covers a broad set of outlets in different languages over more than one year for both comparative and longitudinal studies.

Some of the related datasets publicly available include the News Category Dataset~\citep{misra2022news}, BBC News Archive~\citep{greene2006practical}, AG News~\citep{Zhang2015CharacterlevelCN}, CC News~\citep{Hamborg2017}. However, all of these datasets are mostly or entirely limited to English headlines and/or outlets. Meanwhile, datasets used in projects like~\citet{mazumder2014news} or~\citet{leskovec2009meme} focus on collecting many sources in over a relatively short timespan (see the appendix for a comparison table).

A more global source of news events is offered by the GDELT project~\citep{leetaru2013gdelt}, which collects reports from around the world in a variety of languages. However, the GDELT dataset focuses on tracking events, rather than the news coverage thereof, making it more suitable for event forecasting rather than media coverage or training language models.

\section{Dataset}

\subsection{Collection}

The dataset was collected in three distinct steps. First, using the News API~\citep{newsapi2023}, we gathered the available headlines once a day for each combination of all 54 locations and each of the seven possible categories. Each API call returned a list of about 30-70 article headlines for a total of about 20k instances per day, usually featuring duplicate headlines across locations and categories.

Next, in a pre-processing step, duplicate occurrences of the same article were merged and listed in a list of \texttt{instances} (see below). The author names are anonymized (replacing the names with \texttt{author\#[ID]} where the \texttt{ID} is identical for all articles with matching authors, but distinct otherwise).

Lastly, the final step involved the translation of non-English articles. Using Google Translate~\citep{googletranslate2023}, all articles not originally in English were translated for convenience. Notably, News API appears to only collect articles of a single language for each of the locations, making translation straightforward. Unfortunately, some of the language selections by News API do not seem to fully reflect the local news in a given location (for example Malaysia's articles are all in English), although the headline subjects appear curated for the assigned location.

\subsection{Structure}

\babel{} is structured as a collection of 54 JSON files, one for each location. Each file contains a list of headlines of articles that first occurred in the corresponding location, each of which is represented as a JSON object with the following properties: \texttt{title}, \texttt{description}, \texttt{content}, \texttt{url}, \texttt{urlToImage}, \texttt{publishedAt}, \texttt{author}, \texttt{source}, \texttt{instances}, and \texttt{language}. For articles that are originally in a language other than English, the translated \texttt{title}, \texttt{description}, and \texttt{content} are also included as \texttt{en-title}, \texttt{en-description}, and \texttt{en-content} respectively.

Since each article may appear in multiple locations, categories, or over multiple days, the \texttt{instances} property lists the properties \texttt{location} and \texttt{category} for each instance when the article was collected with timestamp \texttt{collectedAt}. The \texttt{source} property is an object containing the \texttt{id} and \texttt{name} of the news source. The \texttt{language} property is the original language (in ISO 639-1 format) of the article, which is assigned automatically based on the location. A single category is assigned to each instance automatically by News API, and are one of: \texttt{business}, \texttt{entertainment}, \texttt{general}, \texttt{health}, \texttt{science}, \texttt{sports}, or \texttt{technology}.


Notably, the \texttt{content} (and \texttt{en-content}, when present) properties contain nonsense data for articles in languages that use a non-latin alphabet, such as Chinese or Arabic. This is due to a flaw in the News API processing. For details check out the dataset's readme. 

\subsection{Statistics}

In total, we collected a total of 7,419,089 instances of 4,719,199 distinct articles between 8 August 2020 and 29 November 2021, with a breakdown by language in table~\ref{tab:instances_by_language}. More detailed statistics are available in the dataset readme.

\begin{table} 
\centering
\begin{tabular}{lr}
\hline
\textbf{Language} & \textbf{Articles}\\
\hline
English & 1,128,233 \\
Spanish & 455,952 \\
French & 288,328 \\
Chinese & 270,887 \\
German & 259,718 \\
Portuguese & 243,829 \\
Arabic & 178,854 \\
Indonesian & 131,252 \\
Italian & 129,005 \\
Turkish & 122,724 \\
Greek & 119,940 \\
Japanese & 118,475 \\
Polish & 116,904 \\
Russian & 113,395 \\
Dutch & 104,031 \\
Thai & 90,708 \\
Swedish & 86,838 \\
Korean & 83,090 \\
Serbian & 80,040 \\
Hungarian & 73,509 \\
Czech & 70,647 \\
Hebrew & 67,794 \\
Bulgarian & 67,223 \\
Ukrainian & 65,610 \\
Romanian & 54,601 \\
Norwegian & 46,804 \\
Slovak & 43,057 \\
Latvian & 40,006 \\
Lithuanian & 34,719 \\
Slovenian & 33,026 \\
\hline
\end{tabular}
\caption{Number of articles by language.}
\label{tab:instances_by_language}
\end{table}

\section{Analysis}

\begin{figure*}
  \centering
  \includegraphics[width=\linewidth]{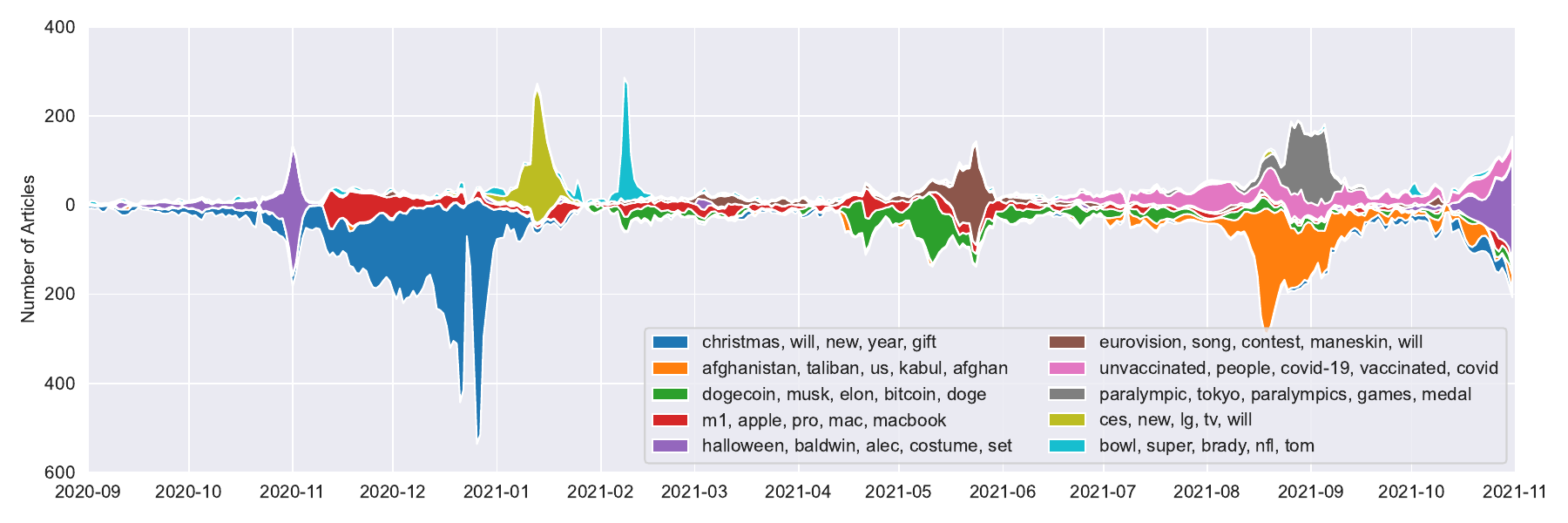}
  \caption{Streamplot showing how many articles appear for some of the most popular events in the dataset, when clustering articles by their titles, with most common tokens for each cluster shown in the legend. Note the qualitative similarity between the news coverage over time of these events and the memes of~\citet{leskovec2009meme}, demonstrating the potential of this dataset for studying the evolution of major news coverage over time across the world.}
  \label{fig:top_articles}
\end{figure*}

A particularly interesting type of analysis enabled by this dataset is the longitudinal comparison of how the same news event is reported in different languages and around the world. To illustrate, let's consider a basic example. We begin by clustering individual articles that discuss the same event. Then, we analyze the distribution and frequency of articles from different countries over time, focusing on that specific event.

To cluster articles that are about the same event, we begin by extracting a bag of words from the article's (English) title where each word is lemmatized as well as removing punctuation and common stopwords (such as "the" or "a"). We use Term Frequency-Inverse Document Frequency (TF-IDF)~\citep{salton1975vector} to define the relevance $R_d$ of each token relative to the other tokens of the articles that occurred on the same day $d$.

\vspace{-4mm}
\begin{equation}
  R_d(w) = \frac{\mathrm{tf}(w, d)}{\sum_{d'} \mathrm{tf}(w, d')} \cdot \log \frac{N}{\mathrm{df}(w)}
\end{equation}

where $\mathrm{tf}(w, d)$ is the number of times the word $w$ occurs in the day $d$, $\mathrm{df}(w)$ is the number of days in which the word $w$ occurs, and $N$ is the total number of days in the dataset.

Using the TF-IDF scores for each word, we define a relevance score $\hat{R}_d(x) = \sum_{w \in x} R_d(w)$ for an article $x$ that occurs first on day $d$ as the sum of the TF-IDF scores of the words in its title. Furthermore, we define a similarity criterion between two articles as the ratio between the sum of all words that occur in both articles weighted by the relevance of each word and the largest relevance score between the two articles.

\vspace{-3mm}
\begin{equation}
  \mathrm{sim}(x, x') = \frac{\sum_{w \in x \cap x'} R_d(w)}{\max(\hat{R}_d(x), \hat{R}_d(x'))}
\end{equation}

If this ratio is greater than some threshold ($=0.25$ in our experiments), we consider the two articles to be in the same group. This means that if the candidate articles have significant overlap between words weighted by how specific those words are to the day.
For the top ten articles with the highest relevance scores every day, we identify all articles in the dataset which, based on our similarity criterion are in the same group to form an event cluster. 

Figure~\ref{fig:top_articles} presents clusters of such articles, identifying the top TF-IDF scores where the clusters are largest—that is, events with the most articles published about them. 

\begin{figure}
  \centering
  \includegraphics[width=\linewidth]{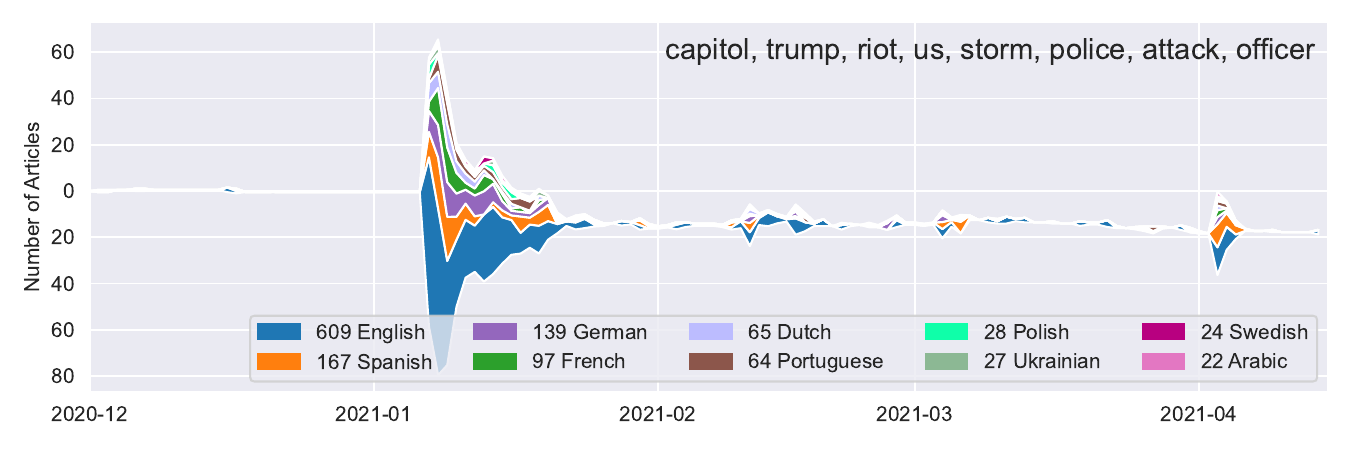}
  \caption{Articles reporting on riots in Washington DC on 6 January 2021. Note how the event is reported in many different languages, but the majority of articles are in English. Additionally, there are several subsequent smaller spikes corresponding to related events, such as the beginning of the formal investigation into the riots.}
  \label{fig:event1}
\end{figure}

\begin{figure}
  \centering
  \includegraphics[width=\linewidth]{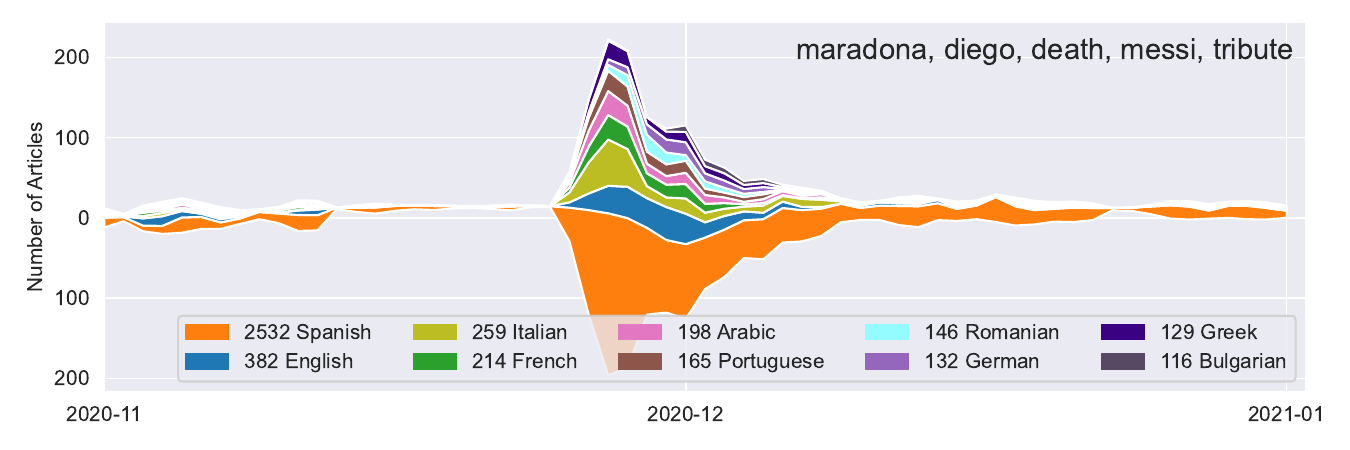}
  \caption{Articles reporting on Diego Maradona's death on 25 November 2020 (and his declining health in the weeks before). Note how in after a few weeks only Spanish articles about the topic continue to appear, underscoring the relative importance of the event in Spanish-speaking countries.}
  \label{fig:event4}
\end{figure}

\begin{figure}
  \centering
  \includegraphics[width=\linewidth]{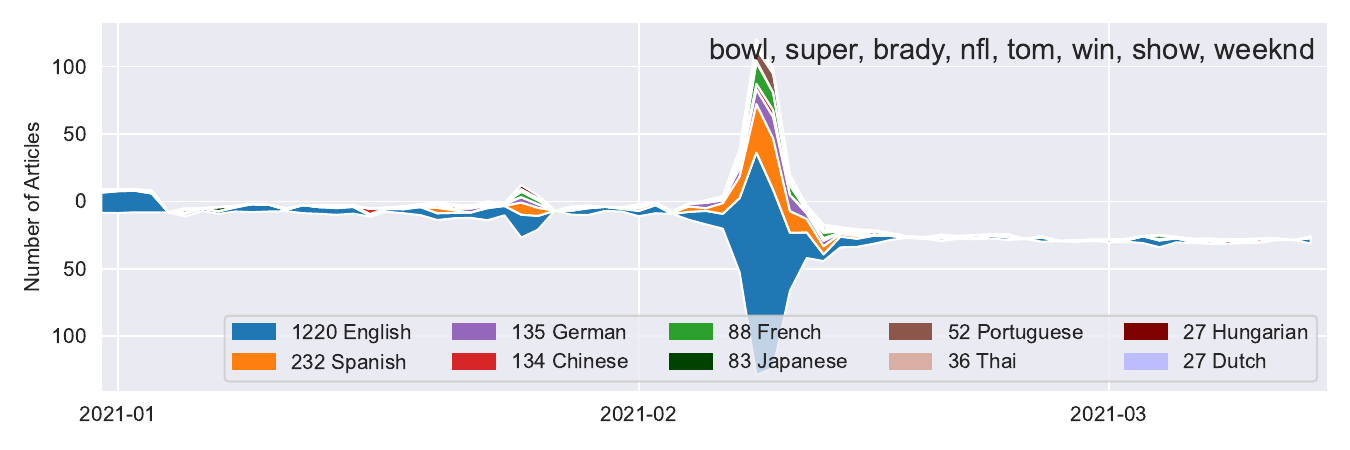}
  \caption{Articles reporting on the Super Bowl on 7 February 2021. Note how unlike unexpected events (such as in figure~\ref{fig:event1}), there is a considerable lead up to the event before the peak, showing the media's anticipation of the event.}
  \label{fig:event2}
\end{figure}

\begin{figure}
  \centering
  \includegraphics[width=\linewidth]{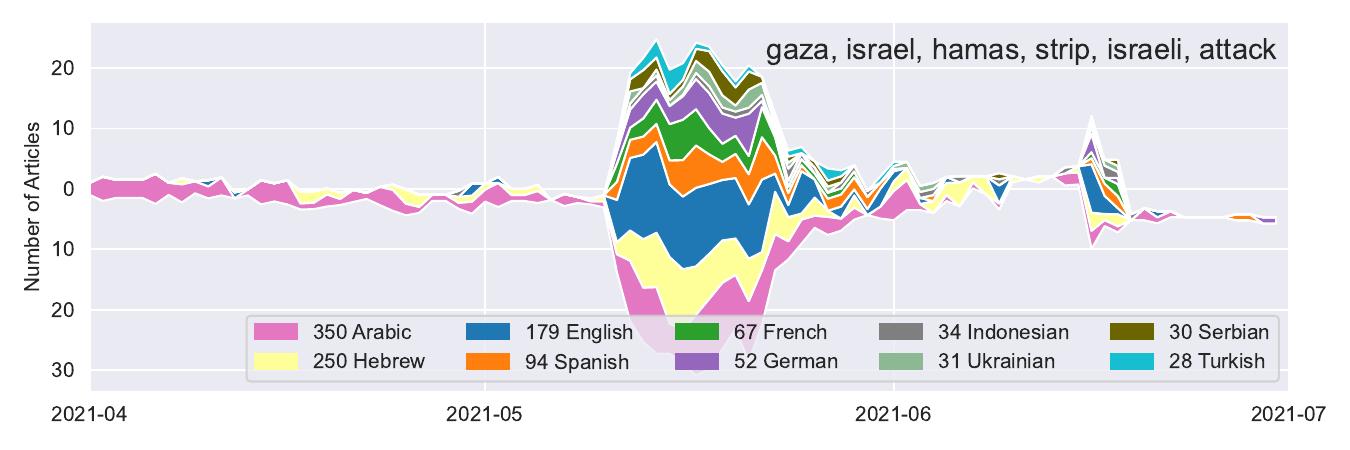}
  \caption{Articles reporting on a crisis between Israel and Gaza on 10 May 2021. Note the prolonged spike for the duration of the crisis, as well as the significant number of articles in Arabic and Hebrew.}
  \label{fig:event3}
\end{figure}

Next, we take a closer look at a few of the largest clusters in figures~\ref{fig:event1}-\ref{fig:event3}. We visualize the \emph{event signatures}, which show how the coverage of the same event varies across different languages by how many articles are published every day. 
For each of the four examples, the plot shows a streamplot breaking down how many articles were published for each of the top ten most common languages for the event as well as the most frequent tokens occurring that cluster in the top right. 
One interesting result from this precursory analysis is a distinct qualitative difference in the event signatures of ``expected'' events (such as in figure~\ref{fig:event2}) compared to ``unexpected'' events (such as in figures~\ref{fig:event1} and~\ref{fig:event3}). For expected events, there is a clear lead-up to the event, with a peak on the day of the event, and a sharp drop-off afterwards. Meanwhile, unexpected events show a sudden spike in coverage, followed by a gradual decline over time. 
This provides a demonstration of the types of analyses that can be conducted with this dataset, offering insights into the diversity and scope of global news coverage.

\section{Conclusion}

In this paper, we introduce a dataset of news headlines from around the world called \babel{}.
The dataset can readily be used for a wide variety of both supervised and unsupervised natural language processing tasks. For example, the included category, location, and language labels can directly be used for article categorization, location classification, or language detection.
However, the dataset also enables more nuanced analyses of global news coverage, such as tracking the evolution of events over time, comparing the coverage of events across different countries and languages, or identifying cultural biases in reporting.

Despite the breadth across languages and time, our dataset is limited to the headlines and short descriptions of news articles. However, URLs to the full articles are included, and since many outlets are incentivized to make their headlines as informative as possible, headlines alone are already a rich source of information for many purposes. Additionally, the dataset is limited to 54 locations, which is a significant improvement over existing datasets which are often limited to a single country or outlet. 
There are some minor issues with the News API, for example that for each location only a single language is represented. We aim to mitigate this issue by collecting headlines directly from the RSS feeds of individual outlets from around the world. However, this may come at the cost of consistency across sources around the world.

In summary, our dataset is a powerful tool for studying the nuances of global news coverage when breaking beyond the language barrier. It provides a simple yet rich foundation for capturing cultural differences in news reporting, offering invaluable data and insights for researchers in the fields of natural language processing, as well as social sciences like media studies or international relations.





\bibliography{anthology,custom}

\newpage

\onecolumn

\appendix

\section{News Headline Dataset Comparison}
\label{sec:appendix}

\begin{table*}[h]
\centering
\begin{tabular}{|l|c|c|c|c|}
\hline
\textbf{Dataset Name}                       & \textbf{Size} & \textbf{Sources} & \textbf{Language/s} & \textbf{Time Span} \\ \hline
\babel{} (ours)                             & 4.7M         & Worldwide & 30 languages        & Aug 2020 - Nov 2021 \\ \hline
News Category Dataset                       & 210k         & HuffPost          & English only        & 2012-2022          \\
\citep{misra2022news}                       &               &                   &                     &                    \\ \hline
BBC News Archive                            & 2225         & BBC               & English only        & 2004-2005          \\
\citep{greene2006practical}                 &               &                   &                     &                    \\ \hline
AG News                                     & 128k         & >2000             & English only        & 2004               \\
\citep{Zhang2015CharacterlevelCN}           &               &                   &                     &                    \\ \hline
CC News                                     & 708k         & Worldwide         & English only        & Jan 2017 - Dec 2019 \\
\citep{Hamborg2017}                         &               &                   &                     &                    \\ \hline
\citet{mazumder2014news} Dataset            & 1.5M         & 87 Indian sources & English only        & Jan - Jun 2014     \\ \hline
\citet{leskovec2009meme} Dataset            & 90M          & US news + blog sites & English only     & Aug - Oct 2008     \\ \hline
GDELT Project                               & >326M        & Worldwide         & >100 Languages      & since 1979         \\
\citep{leetaru2013gdelt}                    &               &                   &                     &                    \\ \hline
\end{tabular}
\caption{Comparison of various existing datasets similar to~\babel{}}
\label{table:dataset_comparison}
\end{table*}




\end{document}